# Composing Finite State Transducers on GPUs

**Arturo Argueta** and **David Chiang**
Department of Computer Science and Engineering
University of Notre Dame
{aargueta,dchiang}@nd.edu

## Abstract

Weighted finite state transducers (FSTs) are frequently used in language processing to handle tasks such as part-of-speech tagging and speech recognition. There has been previous work using multiple CPU cores to accelerate finite state algorithms, but limited attention has been given to parallel graphics processing unit (GPU) implementations. In this paper, we introduce the first (to our knowledge) GPU implementation of the FST composition operation, and we also discuss the optimizations used to achieve the best performance on this architecture. We show that our approach obtains speedups of up to 6× over our serial implementation and 4.5× over OpenFST.

## 1 Introduction

Finite-state transducers (FSTs) and their algorithms (Mohri, 2009) are widely used in speech and language processing for problems such as grapheme-to-phoneme conversion, morphological analysis, part-of-speech tagging, chunking, named entity recognition, and others (Mohri et al., 2002; Mohri, 1997). Hidden Markov models (Baum et al., 1970), conditional random fields (Lafferty et al., 2001) and connectionist temporal classification (Graves et al., 2006) can also be thought of as finite-state transducers.

Composition is one of the most important operations on FSTs, because it allows complex FSTs to be built up from many simpler building blocks, but it is also one of the most expensive. Much work has been done on speeding up composition on a single CPU processor (Pereira and Riley, 1997; Hori and Nakamura, 2005; Dixon et al., 2007; Allauzen and Mohri, 2008; Allauzen et al., 2009; Ladner and Fischer, 1980; Cheng et al., 2007). Methods such as on-the-fly composition, shared data structures, and composition filters have been used to improve time and space efficiency.

There has also been some successful work on speeding up composition using multiple CPU cores (Jurish and Würzner, 2013; Mytkowicz et al., 2014; Jung et al., 2017). This is a challenge because many of the algorithms used in NLP do not parallelize in a straightforward way and previous work using multi-core implementations do not handle the reduction of identical edges generated during the composition. The problem becomes more acute on the graphics processing units (GPUs) architecture, which have thousands of cores but limited memory available. Another problem with the composition algorithm is that techniques used on previous work (such as composition filters and methods to expand or gather transitions using dictionaries or hash tables) do not translate well to the GPU architecture given the hardware limitations and communication overheads.

In this paper, we parallelize the FST composition task across multiple GPU cores. To our knowledge, this is the first successful attempt to do so. Our approach treats the composed FST as a sparse graph and uses some techniques from the work of Merrill et al. (2012); Jung et al. (2017) to explore the graph and generate the composed edges during the search. We obtain a speedup of 4.5× against OpenFST's implementation and 6× against our own serial implementation.

## 2 Finite State Transducers

In this section, we introduce the notation that will be used throughout the paper for the composition task. A weighted FST is a tuple $M = (Q, \Sigma, \Gamma, s, F, \delta)$, where

- $Q$ is a finite set of states.
- $\Sigma$ is a finite input alphabet.
- $\Gamma$ is a finite output alphabet.
- $s \in Q$ is the start state.
- $F \subseteq Q$ are the accept states.
- $\delta : Q \times \Sigma \times \Gamma \times Q \to \mathbb{R}$ is the transition function. If $\delta(q, a, b, r) = p$, we write

$$q \xrightarrow{a:b/p} r.$$

Note that we don't currently allow epsilon transitions; this would require implementation of composition filters (Allauzen et al., 2009), which is not a trivial task on the GPU architecture given the data structures and memory needed. Hence, we leave this for future work.

For the composition task, we are given two weighted FSTs:

$$M_1 = (Q_1, \Sigma, \Gamma, s_1, F_1, \delta_1)$$
$$M_2 = (Q_2, \Gamma, \Delta, s_2, F_2, \delta_2).$$

Call $\Gamma$, the alphabet shared between the two transducers, the *inner* alphabet, and let $m = |\Gamma|$. Call $\Sigma$ and $\Delta$, the input alphabet of $M_1$ and the output alphabet of $M_2$, the *outer* alphabets.

The composition of $M_1$ and $M_2$ is the weighted FST

$$M_1 \circ M_2 = (Q_1 \times Q_2, \Sigma, \Delta, s_1 s_2, F_1 \times F_2, \delta)$$

where

$$\delta(q_1 q_2, a, b, r_1 r_2) = \sum_{b \in \Gamma} \delta_1(q_1, a, c, r_1) \cdot \delta_2(q_2, c, b, r_2).$$

That is, for each pair of transitions with the same inner symbol,

$$q_1 \xrightarrow{a:b/p_1} r_1$$
$$q_2 \xrightarrow{b:c/p_2} r_2,$$

the composed transducer has a transition

$$q_1 q_2 \xrightarrow{a:c/p_1 p_2} r_1 r_2.$$

Transitions with the same source, target, input, and output symbols are merged, adding their weights.

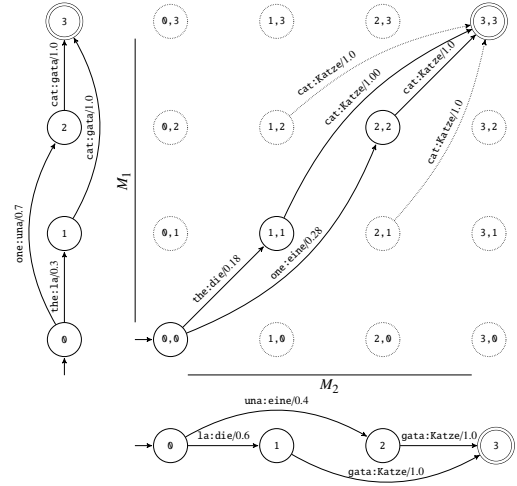

Figure 1: Example composition of two finite state transducers: $M_1$ translates English to Spanish, $M_2$ translates Spanish to German. The center of the image contains the composition of the two input transducers. This new transducer translates English to German. The dotted states and transitions are those that cannot be reached from the start state.

## 3 Method

In this section, we describe our composition method and its implementation.

### 3.1 Motivation

If implemented naïvely, the above operation is inefficient. Even if $M_1$ and $M_2$ are trim (have no states that are unreachable from the start state or cannot reach the accept state), their composition may have many unreachable states. Figure 1 shows a clear example where the transducers used for composition are trim, yet several states (drawn as dotted circles) on the output transducers cannot be reached from the start state. The example also shows composed transitions that originate from unreachable states. As a result, a large amount of time and memory may be spent creating states and composing transitions that will not be reachable nor needed in practice. One solution to avoid the problem is to compose only the edges and states that are reachable from the start state on the output transducer to avoid unnecessary computations and reduce the overall memory footprint.

We expect this problem to be more serious when the FSTs to be composed are sparse, that is, when there are many pairs of states without a transition between them. And we expect that FSTs used in natural language processing, whether they are constructed by hand or induced from data, will often be sparse.

| | Data | Transitions | Nonzero | % Nonzero |
|---|---|---|---|---|
| 1k | de-en | 21.7M | 16.5k | 0.076% |
| | en-de | 12.3M | 15.4k | 0.125% |
| | en-es | 12.5M | 15.5k | 0.124% |
| | en-it | 13.1M | 16.3k | 0.124% |
| 10k | de-en | 394M | 114k | 0.029% |
| | en-de | 138M | 93.9k | 0.067% |
| | en-es | 135M | 93.3k | 0.068% |
| | en-it | 143M | 97.2k | 0.067% |
| 15k | de-en | 634M | 158k | 0.025% |
| | en-de | 201M | 126k | 0.062% |
| | en-es | 195M | 125k | 0.064% |
| | en-it | 209M | 131k | 0.062% |

Table 1: FSTs used in our experiments. Key: Data = language pair used to generate the transducers; Transitions = maximum possible number of transitions; Nonzero = number of transitions with nonzero weight; % Nonzero = percent of possible transitions with nonzero weight. The left column indicates the number of parallel sentences used to generate the transducers used for testing.

For example, below (Section 4.1), we will describe some FSTs induced from parallel text that we will use in our experiments. We measured the sparsity of these FSTs, shown in Table 1. These FSTs contain very few non-zero connections between their states, suggesting that the output of the composition will have a large number of unreachable states and transitions. The percentage of non-zero transitions found in the transducers used for testing decreases as the transducer gets larger. Therefore, when composing FSTs, we want to construct only reachable states, using a traversal scheme similar to breadth-first search to avoid the storage and computation of irrelevant elements.

### 3.2 Serial composition

We first present a serial composition algorithm (Algorithm 2). This algorithm performs a breadth-first search (BFS) of the composed FST beginning from the start state, so as to avoid creating inaccessible states. As is standard, the BFS uses two data structures, a frontier queue ($A$) and a visited set ($Q$), which is always a superset of $A$. For each state $q_1 q_2$ popped from $A$, the algorithm composes

**Algorithm 1** Serial composition algorithm.
**Input** Transducers: $M_1 = (Q_1, \Sigma, \Gamma, s_1, F_1, \delta_1)$
$M_2 = (Q_2, \Gamma, \Delta, s_2, F_2, \delta_2)$
**Output** Transducer: $M_1 \circ M_2$

1: $A \leftarrow \{s_1 s_2\}$ ▷ Queue of states to process
2: $Q \leftarrow \{s_1 s_2\}$ ▷ Set of states created so far
3: $\delta \leftarrow \emptyset$ ▷ Transition function
4: **while** $|A| > 0$ **do**
5:     $q_1 q_2 \leftarrow \text{pop}(A)$
6:     **for** $q_1 \xrightarrow{a:b/p_1} r_1 \in \delta_1$ **do**
7:        **for** $q_2 \xrightarrow{b:c/p_2} r_2 \in \delta_2$ **do**
8:          $\delta(q_1 q_2, a, c, r_1 r_2) \mathrel{+}= p_1 p_2$
9:          **if** $r_1 r_2 \notin Q$ **then**
10:            $Q \leftarrow Q \cup \{r_1 r_2\}$
11:            $\text{push}(A, r_1 r_2)$
12: **return** $(Q, \Sigma, \Delta, s_1 s_2, F_1 \times F_2, \delta)$

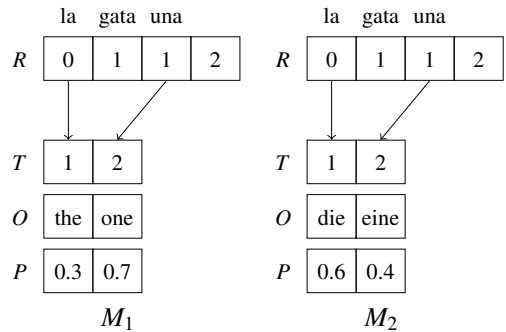

Figure 2: Example CSR-like representation of state 0 for transducers $M_1$ and $M_2$ from Figure 1.

all transitions from $q_1$ with all transitions from $q_2$ that have the same inner symbol. The composed edges are added to the final transducer, and the corresponding target states $q_1 q_2$ are pushed into $A$ for future expansion. The search finishes once $A$ runs out of states to expand.

### 3.3 Transducer representation

Our GPU implementation stores FST transition functions in a format similar to compressed sparse row (CSR) format, as introduced by our previous work Argueta and Chiang (2017). For the composition task we use a slightly different representation. An example of the adaptation is shown in Figure 2. The transition function $\delta$ for the result is stored in a similar fashion. The storage method is defined as follows:

- $z$ is the number of transitions with nonzero weight.

- $R$ is an array of length $|Q|m + 1$ containing offsets into the arrays $T$, $O$, and $P$. If the states are numbered $0, \ldots, |Q| - 1$ and the inner symbols are numbered $0, \ldots m - 1$, then state $q$'s outgoing transitions on inner symbol $b$ can be found starting at the offset stored in $R[qm + b]$. The last offset index, $R[|Q|m + 1]$, must equal $z$.

- $T[k]$ is the target state of the $k$th transition.

- $O[k]$ is the outer symbol of the $k$th transition.

- $P[k]$ is the weight of the $k$th transition.

Similarly to several toolkits (such as OpenFST), we require the edges in $T, O, P$ to be sorted by their inner symbols before executing the algorithm, which allows faster indexing and simpler parallelization.

### 3.4 Parallel composition

Our parallel composition implementation has the same overall structure as the serial algorithm, and is shown in Algorithm 2. The two transducers to be composed are stored on the GPU in global memory, in the format described in Section 3.3. Both transducers are sorted according to their inner symbol on the CPU and copied to the device. The memory requirements for a large transducer complicates the storage of the result on the GPU global memory. If the memory of states and edges generated by both inputs does not fit on the GPU, then the composition cannot be computed using only device memory. The execution time will also be affected if the result lives on the device and there is a limited amount of memory available for temporary variables created during the execution. Therefore, the output transducer must be stored on the host using page-locked memory, with the edge transitions unsorted.

Page-locked, or pinned, memory is memory that will not get paged out by the operating system. Since this memory cannot be paged out, the amount of RAM available to other applications will be reduced. This enables the GPU to access the host memory quickly. Pinned memory provides better transfer speeds since the GPU creates different mappings to speed up `cudaMemcpy` operations on host memory. Allocating pinned memory consumes more time than a regular `malloc`,

**Algorithm 2** Parallel composition algorithm.
**Input** Transducers: $M_1 = (Q_1, \Sigma, \Gamma, s_1, F_1, \delta_1)$
$M_2 = (Q_2, \Gamma, \Delta, s_2, F_2, \delta_2)$
**Output** Transducer: $M_1 \circ M_2$

1: $A \leftarrow \{s_1 s_2\}$ ▷ Queue of states to process
2: $Q \leftarrow \{s_1 s_2\}$ ▷ Set of states visited
3: $\delta \leftarrow []$ ▷ List of transitions
4: **while** $|A| > 0$ **do**
5: $\quad q_1 q_2 \leftarrow \text{pop}(A)$
6: $\quad \delta_d \leftarrow []$
7: $\quad A_d \leftarrow \emptyset$
8: $\quad H \leftarrow \emptyset$
9: $\quad red \leftarrow \text{false}$
10: $\quad$ **parfor** $b \in \Gamma$ **do** ▷ kernels
11: $\quad\quad$ **parfor** $q_1 \xrightarrow{a:b/p_1} r_1$ **do** ▷ threads
12: $\quad\quad\quad$ **parfor** $q_2 \xrightarrow{b:c/p_2} r_2$ **do** ▷ threads
13: $\quad\quad\quad\quad$ append $q_1 q_2 \xrightarrow{a:c/p_1 p_2} r_1 r_2$ to $\delta_d$
14: $\quad\quad\quad\quad$ **if** $h(a, c, r_1 r_2) \in H$ **then**
15: $\quad\quad\quad\quad\quad$ $red \leftarrow \text{true}$
16: $\quad\quad\quad\quad$ **else**
17: $\quad\quad\quad\quad\quad$ add $h(a, c, r_1 r_2)$ to $H$
18: $\quad\quad\quad\quad$ **if** $r_1 r_2 \notin Q$ **then**
19: $\quad\quad\quad\quad\quad$ $A_d \leftarrow A_d \cup \{r_1 r_2\}$
20: $\quad\quad\quad\quad\quad$ $Q \leftarrow Q \cup \{r_1 r_2\}$
21: $\quad$ concatenate $\delta_d$ to $\delta$
22: $\quad$ **for** $q \in A_d$ **do** push$(A, q)$
23: $\quad$ **if** $red$ **then**
24: $\quad\quad$ sort $\delta[q_1 q_2]$
25: $\quad\quad$ reduce $\delta[q_1 q_2]$
26: **return** $(Q, \Sigma, \Delta, s_1 s_2, F_1 \times F_2, \delta)$

therefore it should be done sporadically. In this work, pinned memory is allocated only once at start time and released once the composition has been completed. Using page-locked memory on the host side as well as pre-allocating memory on the device decreases the time to both copy the results back from the GPU, and the time to reuse device structures used on different kernel methods.

**Generating transitions** The frontier queue $A$ is stored on the host. For each state $q_1 q_2$ popped from $A$, we need to compose all outgoing transitions of $q_1$ and $q_2$ obtained from $M_1$ and $M_2$ respectively. Following previous work (Merrill et al., 2012; Jurish and Würzner, 2013), we create these in parallel, using the three **parfor** loops in lines 10–12. Although these three loops are written the same way in pseudocode for simplic-ity, in actuality they use two different parallelization schemes in the actual implementation of the algorithm.

The outer loop launches a CUDA kernel for each inner symbol $b \in \Gamma$. For example, to compose the start states in Figure 1, three kernels will be launched (one for *la*, *gata*, and *una*). Each of these kernels composes all outgoing transitions of $q_1$ with output $b$ with all outgoing transitions of $q_2$ with input $b$. Each of these kernels is executed in a unique stream, so that a higher parallelization can be achieved. Streams are used in CUDA programming to minimize the number of idle cores during execution. A stream is a group of commands that execute in-order on the GPU. What makes streams useful in CUDA is the ability to execute several asynchronously. If more than one kernel can run asynchronously without any interdependence, the assignment of kernel calls to different streams will allow a higher speedup by minimizing the amount of idling cores during execution. All kernel calls using streams are asynchronous to the host making synchronization between several different streams necessary if there exist data dependencies between different parts of the execution pipeline. Asynchronous memory transactions can also benefit from streams, if these operations do not have any data dependencies.

We choose a kernel block size of 32 for the kernel calls since this is the amount of threads that run in parallel on all GPU streaming multiprocessors at any given time. If the number of threads required to compose a tuple of states is not divisible by 32, the number of threads is rounded up to the closest multiple. When several input tuples generate less than 32 edges, multiple cores will remain idle during execution. Our approach obtains better speedups when the input transducers are able to generate a large amount of edges for each symbol $b$ and each state tuple on the result. In general, the kernels may take widely varying lengths of time based on the amount of composed edges; using streams enables the scheduler to minimize the number of idle cores.

The two inner loops represent the threads of the kernel; each composes a pair of transitions sharing an inner symbol $b$. Because these transitions are stored contiguously (Figure 2), the reads can be coalesced, meaning that the memory reads from the parallel threads can be combined into one transaction for greater efficiency. Figure 2 shows how the edges for a transducer are stored in global memory to achieve coalesced memory operations each time the edges of a symbol $b$ associated with a state tuple $q_1,q_2$ need to be composed.

Figure 2 shows how the edges leaving the start state tuple for transducers $M_1$ and $M_2$ are stored. As mentioned above, three kernels will be launched to compose the transitions leaving the start states, but only two will be executed (because there are no transitions on *gata* for both start states). For $R[\texttt{la}]$ on machine $M_1$, only one edge can output la given $R[\texttt{la} + 1] - R[\texttt{la}] = 1$, and machine $M_2$ has one edge that reads *la* given $R[\texttt{la} + 1] - R[\texttt{la}] = 1$. For this example, $R[\texttt{la}]$ points to index 0 on $T, O, P$ for both states. This means that only one edge will be generated from the composition $(0, 0 \xrightarrow{\texttt{the:die}/0.18} 1, 1)$. For symbol *gata*, no edges can be composed given $R[\texttt{gata} + 1] - R[\texttt{gata}] = 0$ on both machines, meaning that no edges read or output that symbol. Finally, for $R[\texttt{una}]$ on machine $M_1$ and $M_2$, one edge can be generated $(0, 0 \xrightarrow{\texttt{one:eine}/0.28} 2, 2)$ given the offsets in $R$ for both input FSTs. If $n_1$ edges can be composed for a symbol $b$ on one machine and $n_2$ from the other one, the kernel will generate $n_1 n_2$ edges.

The composed transitions are first appended to a pre-allocated buffer $\delta_d$ on the GPU. After processing the valid compositions leaving $q_1 q_2$, all the transitions added in $\delta_d$ are appended in bulk to $\delta$ on the host.

**Updating frontier and visited set**  Each destination state $r_1 r_2$, if previously unvisited, needs to be added to both $A$ and $Q$. Instead of adding it directly to $A$ (which is stored on the host), we add it to a buffer $A_d$ stored on the device to minimize the communication overhead between the host and the device. After processing $q_1 q_2$ and synchronizing all streams, $A_d$ is appended in bulk to $A$ using a single `cudaMemcpy` operation.

The visited set $Q$ is stored on the GPU device as a lookup table of length $|Q_1||Q_2|$. Merrill et al. (2012) perform BFS using two stages to obtain the states and edges needed for future expansion. Similarly, our method performs the edge expansion using two steps by using the lookup table $Q$. The first step of the kernel updates $Q$ and all visited states that need to be added to $A_d$. The second step appends all the composed edges to $\delta$ in parallel. Since several threads check the table in parallel, an atomic operation (`atomicOr`) is used to check and update each value on the table in a consistent fashion. $Q$ also functions as a map to convert the state tuple $q_1 q_2$ into a single integer. Each time a tuple is not in $Q$, the structure gets updated with the total number of states generated plus one for a specific pair of states.

**Reduction** Composed edges with the same source, target, input, and output labels must be merged, summing their probabilities. This is done in lines 23–25, which first sort the transitions and then merge and sum them. To do this, we pack the transitions into an array of keys and an array of values. Each key is a tuple $(a, c, r_1 r_2)$ packed into a 64-bit integer. We then use the sort-by-key and reduce-by-key operations provided by the Thrust library. The mapping of tuples to integers is required for the sort operation since the comparisons required for the sorting can be made faster than using custom data structures with a custom comparison operator. [1]

Because the above reduction step is rather expensive, lines 14–17 use a heuristic to avoid it if possible. $H$ is a set of transitions represented as a hash table without collision resolution, so that lookups can yield false positives. If *red* is false, then there were no collisions, so the reduction step can be skipped. The hash function is simply $h(a, c, r_1 r_2) = a + c|\Sigma|$. In more detail, $H$ actually maps from hashes to integers. Clearing $H$ (line 8) actually just increments a counter $i$; storing a hash $k$ is implemented as $H[k] \leftarrow i$, so we can test whether $k$ is a member by testing whether $H[k] = i$. An atomic operation (`atomicExch`) is used to consistently check $H$ since several threads update this variable asynchronously.

## 4 Experiments

We tested the performance of our implementation by constructing several FSTs of varying sizes and comparing our implementation against other baselines.

### 4.1 Setup

In our previous work (Argueta and Chiang, 2017), we created transducers for a toy translation task. We trained a bigram language model (as in Figure 3a) and a one-state translation model (as in Figure 3) with probabilities estimated from GIZA++ Viterbi word alignments. Both were trained on the Europarl corpus. We then pre-composed them using the Carmel toolkit (Graehl, 1997).

---
[1] https://thrust.github.io/

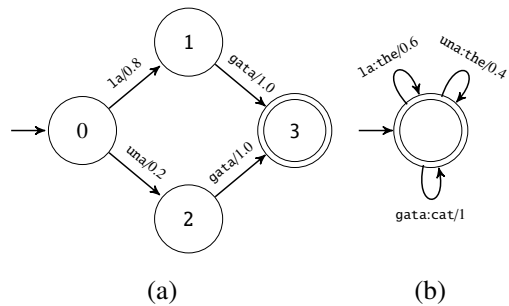

Figure 3: The transducers used for testing were obtained by pre-composing: (a) a language model and (b) a translation model. These two composed together form a transducer that can translate an input sequence from one language (here, Spanish) into another language (here, English).

We used the resulting FSTs to test our parallel composition algorithm, composing a German-to-English transducer with a English-to-*t* transducer to translate German to language *t*, where *t* is German, Spanish, or Italian.

Our experiments were tested using two different architectures. The serial code was measured using a 16-core Intel Xeon CPU E5-2650 v2, and the parallel implementation was executed on a system with a GeForce GTX 1080 Ti GPU connected to a 24-core Intel Xeon E5-2650 v4 processor.

### 4.2 Baselines

In this work, OpenFST (Allauzen et al., 2007) and our serial implementation (Algorithm 1) were used as a baseline for comparison. OpenFST is a toolkit developed by Google as a successor of the AT&T Finite State Machine library. For consistency, all implementations use the OpenFST text file format to read and process the transducers.

### 4.3 Results

OpenFST's composition operation can potentially create multiple transitions (that is, two or more transitions with the same source state, destination state, input label, and output label); a separate function (`ArcSumMapper`) must be applied to merge multiple transitions and sum their weights. Previous work also requires an additional step if identical edges need to be merged. For this reason, we compare our implementation against OpenFST both with and without the reduction of transitions with an identical source,target,input, and output. We analyzed the time to compose all possible edges without performing any reductions (Algorithm 1, line 8). The second setup analyzes the time it takes to compute the composition and the arc summing of identical edges generated during the process.

|        |                  |        | Training size (lines) | | | | | |
|        |                  |        | 1000 | | 10000 | | 15000 | |
| Method | Hardware | Target | Time | Ratio | Time | Ratio | Time | Ratio |
|---|---|---|---|---|---|---|---|---|
| OpenFST | Xeon E5 | DE | 0.52 | 0.78 | 69.51 | 3.56 | 157.16 | 4.38 |
| our serial | Xeon E5 | DE | **0.21** | **0.31** | 28.47 | 1.45 | 72.33 | 2.02 |
| our parallel | GeForce GTX 1080 | DE | 0.67 | 1.00 | **19.54** | **1.00** | **35.89** | **1.00** |
| OpenFST | Xeon E5 | ES | 0.46 | 0.72 | 55.62 | 2.97 | 137.16 | 4.07 |
| our serial | Xeon E5 | ES | **0.19** | **0.30** | 23.30 | 1.24 | 62.42 | 1.85 |
| our parallel | GeForce GTX 1080 | ES | 0.64 | 1.00 | **18.72** | **1.00** | **33.71** | **1.00** |
| OpenFST | Xeon E5 | IT | 0.54 | 0.79 | 60.66 | 3.05 | 136.06 | 3.91 |
| our serial | Xeon E5 | IT | **0.21** | **0.31** | 25.58 | 1.28 | 119.84 | 3.45 |
| our parallel | GeForce GTX 1080 | IT | 0.68 | 1.00 | **19.88** | **1.00** | **34.76** | **1.00** |

Table 2: This table shows how the total running time of our GPU implementation compares against all other methods. Times (in seconds) are for composing two transducers using English as the shared input/output vocabulary and German as the source language of the first transducer (de-en,en-*). Ratios are relative to our parallel algorithm on the GeForce GTX 1080 Ti.

Table 2 shows the performance of the parallel implementation and the baselines without reducing identical edges. For the smallest transducers, our parallel implementation is slower than the baselines (0.72× compared to OpenFST and 0.30× compared to our serial version). With larger transducers, the speedups increase up to 4.38× against OpenFST and 2.02× against our serial implementation. Larger speedups are obtained for larger transducers because the GPU can utilize the streaming multiprocessors more fully. On the other hand, the overhead created by CUDA calls, device synchronization, and memory transfers between the host CPU and the device might be too expensive when the inputs are too small.

Table 3 shows the performance of all implementations with the reduction operation. Again, for the smallest transducers we can see a similar behavior, our parallel implementation is slower (0.30× against OpenFST and 0.39× against our serial version). Speedups improve with the larger transducers, eventually achieving a 4.52× speedup over OpenFST and a 6.26× speedup over our serial implementation of the composition algorithm.

### 4.4 Discussion

One comparison missing above is a comparison against a multicore processor. We attempted to compare against a parallel implementation using OpenMP on a single 16-core processor, but it did not yield any meaningful speedup, and even slowdowns of up to 10%. We think the reason for this is that because the BFS-like traversal of the FST makes it impractical to process states in parallel, the best strategy is to process and compose transitions in parallel. This very fine-grained parallelism does not seem suitable for OpenMP, as the overhead due to thread initialization and synchronization is higher than the time to execute the parallel sections of the code where the actual composition is calculated. According to our measurements, the average time to compose two transitions is 7.4 nanoseconds, while the average time to create an OpenMP thread is 10.2 nanoseconds. By contrast, the overhead for creating a CUDA thread seems to be around 0.4 nanoseconds. While a different parallelization strategy may exist for multicore architectures, at present, our finding is that GPUs, or other architectures with a low cost to create and destroy threads, are much more suitable for the fine grained operations used for the composition task.

|        |                 |        | Training size (lines) | | | | | |
|--------|-----------------|--------|------|------|--------|------|--------|------|
|        |                 |        | 1000 | | 10000 | | 15000 | |
| Method | Hardware        | Target | Time | Ratio | Time | Ratio | Time | Ratio |
| OpenFST     | Xeon E5          | DE | **0.87** | **0.41** | 148.11 | 3.19 | 374.72 | 4.52 |
| our serial  | Xeon E5          | DE | 0.96 | 0.45 | 213.27 | 4.59 | 518.97 | 6.26 |
| our parallel | GeForce GTX 1080 | DE | 2.11 | 1.00 | **47.70** | **1.00** | **82.88** | **1.00** |
| OpenFST     | Xeon E5          | ES | **0.60** | **0.30** | 116.45 | 2.66 | 279.85 | 3.57 |
| our serial  | Xeon E5          | ES | 0.77 | 0.39 | 202.15 | 4.61 | 390.29 | 4.97 |
| our parallel | GeForce GTX 1080 | ES | 2.00 | 1.00 | **45.30** | **1.00** | **78.38** | **1.00** |
| OpenFST     | Xeon E5          | IT | **0.76** | **0.36** | 130.61 | 2.87 | 309.28 | 3.79 |
| our serial  | Xeon E5          | IT | 1.06 | 0.50 | 158.57 | 3.48 | 427.51 | 5.24 |
| our parallel | GeForce GTX 1080 | IT | 2.12 | 1.00 | **47.04** | **1.00** | **81.54** | **1.00** |

Table 3: This table shows how the total running time of our GPU implementation compares against all other methods. Times (in seconds) are for composing two transducers and performing edge reduction using English as the shared input/output vocabulary and German as the source language of the first transducer (de-en,en-*). Ratios are relative to our parallel algorithm on the GeForce GTX 1080 Ti.

## 5 Future Work

For future work, other potential bottlenecks could be addressed. The largest bottleneck is the queue used on the host to keep track of the edges to expand on the GPU. Using a similar data structure on the GPU to keep track of the states to expand would yield higher speedups. The only challenge of using such a data structure is the memory consumption on the GPU. If the two input transducers contain a large number of states and transitions, the amount of memory needed to track all the states and edges generated will grow significantly. Previous work (Harish and Narayanan, 2007) has shown that state queues on the GPU cause a large memory overhead. Therefore, if state expansion is moved to the GPU, the structures used to keep track of the states must be compressed or occupy the least amount of memory possible on the device in order to allocate all structures required on the device. The queue will also require a mechanism to avoid inserting duplicate tuples into the queue.

For the reduction step, speedups can be achieved if the sort and reduce operations can be merged with the edge expansion part of the method. The challenge of merging identical edges during expansion is the auxiliary memory that will be required to store and index intermediate probabilities. It can be doable if the transducers used for the composition are small. In that case, the reduce operation might not yield significant speedups given the fact that the overhead to compose small transducers is too high when using a GPU architecture.

## 6 Conclusion

This is the first work, to our knowledge, to deliver a parallel GPU implementation of the FST composition algorithm. We were able to obtain speedups of up to 4.5× over a serial OpenFST baseline and 6× over the serial implementation of our method. This parallel method considers several factors, such as host to device communication using page-locked memory, storage formats on the device, thread configuration, duplicate edge detection, and duplicate edge reduction. Our implementation is available as open-source software.[2]

### Acknowledgements

We thank the anonymous reviewers for their helpful comments. This research was supported in part by an Amazon Academic Research Award and a hardware grant from NVIDIA.

---

[2] https://bitbucket.org/aargueta2/parallel_composition